\documentclass[11pt,a4paper]{article}
\usepackage[hyperref]{acl2020}
\usepackage{times}
\usepackage{latexsym}

\usepackage{amsmath,amsfonts,bm}

\def\eqref#1{equation~\ref{#1}}

\def\1{\bm{1}}

\def\vb{{\bm{b}}}

\def\vf{{\bm{f}}}

\def\vv{{\bm{v}}}

\def\vx{{\bm{x}}}

\def\mW{{\bm{W}}}

\DeclareMathAlphabet{\mathsfit}{\encodingdefault}{\sfdefault}{m}{sl}
\SetMathAlphabet{\mathsfit}{bold}{\encodingdefault}{\sfdefault}{bx}{n}

\newcommand{\BERT}{\mathrm{BERT}}

\usepackage{graphicx}
\usepackage{algorithm}
\usepackage{algpseudocode}
\usepackage{booktabs}
\usepackage{marvosym}
\usepackage{multirow}
\usepackage{graphicx}

\aclfinalcopy 

\newcommand{\BLEURT}{\textsc{Bleurt}}

\newcommand{\En}{\texttt{En}}

\newcommand{\Gu}{\texttt{Gu}}
\newcommand{\Kk}{\texttt{Kk}}
\newcommand{\Lt}{\texttt{Lt}}
\newcommand{\Ru}{\texttt{Ru}}
\newcommand{\Zh}{\texttt{Zh}}

\newcommand{\X}{\texttt{X}}
\newcommand{\Y}{\texttt{Y}}

\newcommand{\arkcomment}[3]{\ensuretext{\textcolor{#3}{[#1 #2]}}}
\renewcommand{\arkcomment}[3]{} 

\title{Learning to Evaluate Translation Beyond English\\
\large \BLEURT{} Submissions to the WMT Metrics 2020 Shared Task}

\author{Thibault Sellam \ \ \ Amy Pu\thanks{ \, Work done during a summer internship. Permanent email address: \texttt{amy\_pu@brown.edu}.} \ \ \ Hyung Won Chung\thanks{ \, Work done as a member of the Google AI Residency Program.} \ \ \ Sebastian Gehrmann\\
        \texttt{\small \{tsellam, puamy, hwchung, gehrmann\}@google.com} 
        \AND
       Qijun Tan \ \ \ Markus Freitag \ \ \ Dipanjan Das \ \ \ Ankur P. Parikh\\
        
        \texttt{\small \{qijuntan, freitag, dipanjand, aparikh\}@google.com} \\
        Google Research 
}

\begin{document}

\maketitle

\begin{abstract}
The quality of machine translation systems has dramatically improved over the last decade, and as a result, evaluation has become an increasingly challenging problem. This paper describes our contribution to the WMT 2020 Metrics Shared Task, the main benchmark for automatic evaluation of translation. We make several submissions based on \BLEURT{}, a previously published metric which uses transfer learning. We extend the metric beyond English and evaluate it on 14 language pairs for which fine-tuning data is available, as well as 4 ``zero-shot'' language pairs, for which we have no labelled examples. Additionally, we focus on English to German and demonstrate how to combine \BLEURT{}'s predictions with those of \textsc{YiSi} and use alternative reference translations to enhance the performance. Empirical results show that the models achieve competitive results on the WMT Metrics 2019 Shared Task, indicating their promise for the 2020 edition.
\end{abstract}

\section{Introduction}
The recent progress in machine translation models has led researchers to question the use of n-gram overlap metrics such as \textsc{BLEU}, which focus solely on surface-level aspects of the generated text, and thus may correlate poorly with human evaluation~\cite{papineni2002bleu,lin2004rouge,ma2019results,mathur2020tangled,belz2006comparing,callison2006re}.
This has led to a surge of interest for more flexible metrics that use machine learning to capture semantic-level information~\cite{celikyilmaz2020evaluation}.
Popular examples of such metrics include
\textsc{YiSi-1}~\cite{lo2019yisi},
\textsc{ESIM}~\cite{mathur2019putting},
\textsc{BERTscore}~\cite{zhang2019BERTscore},
the Sentence Mover's Similarity~\cite{zhao2019moverscore,clark2019sentence},
and \BLEURT{}~\cite{sellam2020bleurt}. 
These metrics utilize contextual embeddings from large models such as \textsc{BERT}~\cite{devlin2018bert} which have been shown to capture linguistic information beyond surface-level aspects~\cite{tenney2019bert}.

The WMT Metrics 2020 Shared Task is the reference benchmark for evaluating these metrics in the context of machine translation. It tests the evaluation of systems that are to-English ($\X\rightarrow \En$) and to other languages ($\X \rightarrow \Y$), which requires a multilingual approach. An additional challenge for learned metrics is that human ratings are not available for all language pairs, and therefore, the models must use unlabeled data and perform zero-shot generalization. 

We describe several learned metrics based on \BLEURT{}~\citep{sellam2020bleurt}, originally developed for English data. We first extend \BLEURT{} to the multilingual setup, and show that our approach achieves competitive results on the WMT Metrics 2019 Shared Task.\footnote{We use the following languages for fine-tuning and/or testing: Chinese, Czech, German, English, Estonian, Finnish, French, Gujarati, Kazakh, Lithuanian, Russian, and Turkish. In addition, we also pre-train on Inuktitut, Japanese, Khmer, Pastho, Polish, Romanian, and Tamil.} We also present several simple \textsc{BERT}-based baselines, which we submit for analysis.
Finally, we focus on English to German and enhance \BLEURT{}'s performance by combining its predictions with those of \textsc{YiSi}~\cite{lo2019yisi} as well as by using alternative  references.

\section{Background and Notations}

\paragraph{Task}
Reference-based NLG evaluation seeks to assign a score to a triplet of sentences \emph{(input, reference, candidate)}, where  \emph{input} is a sentence in the source language, \emph{reference} is a reference translation kept secret at inference time, and \emph{candidate} is a translation produced by an MT system.

Similar to \textsc{BLEU}~\cite{papineni2002bleu} and the previous editions of the WMT Metrics shared task, we omit the input and treat the task as a regression problem : we aim to learn a function $\vf : (\vx, \tilde{\vx}) \rightarrow y$ that predicts a quality score $y$ for a candidate sentence $\tilde{\vx} = (\tilde{x}_1,..,\tilde{x}_{p})$ given a reference sentence $\vx=(x_1,..,x_q)$. The function is supervised on a corpus of $N$ human ratings $\{(\vx_i, \tilde{\vx}_i, y_i)\}_{n=1}^{N}$.

\paragraph{\BLEURT{}}
Most experiments presented in this paper are based on \BLEURT{}, a metric that leverages transfer learning to achieve high accuracy and increase robustness~\cite{sellam2020bleurt}. \BLEURT{} is a \textsc{BERT}-based regression model~\cite{devlin2018bert}. It embeds sentence pairs into a fixed-width vector $\vv_{\textrm{BERT}} =\BERT(\vx, \tilde{\vx})$
with a pre-trained Transformer, and feeds this vector to a linear layer: 
$$
\hat{y} = \vf(\vx, \tilde{\vx}) = \mW \vv_{\textrm{BERT}} + \vb 
$$
where $\mW$ and $\vb$ are the weight matrix and bias vector respectively.

In its original (English) version, \BLEURT{} is trained in three stages. (1) It is initialized from a publicly available \textsc{BERT} checkpoint. (2) The model is then ``warmed up'' by exposing it to millions of sentence pairs $(\vx, \tilde{\vx})$, obtained by randomly perturbing sentences from Wikipedia. During this phase, the model learns to predict a wide range of similarity scores that include existing metrics (\textsc{BERTscore}, \textsc{BLEU}, \textsc{ROUGE}), scores from an entailment model, and the likelihood that $\tilde{\vx}$ was generated from $\vx$ with a round-trip translation by a given translation model. We denote this stage as \emph{mid-training}. (3) In the final stage, the model is fine-tuned on human ratings from WMT Metrics~\cite{ma2017results, ma2018results, ma2019results}, using a regression loss $\ell_{\textrm{supervised}} = \frac{1}{N} \sum_{n=1}^{N} \|y_i  - \hat{y} \|^2 $. We found that English \BLEURT{} achieved competitive performance on four academic datasets, WebNLG~\cite{gardent2017webnlg}, and the WMT Metrics Shared Task years 2017 to 2019.

\section{Extending BLEURT Beyond English}



\subsection{Modeling}
An approach to extend BLEURT would be to use \textsc{mBERT}, the public version of \textsc{BERT} pre-trained on 104 languages, and ``mid-train'' with non-English signals as described above. Yet, the evidence we gathered from early experiments were inconclusive. On the other hand, we did observe that models trained on several languages were often more accurate than monolingual models, possibly due to the larger amount of fine-tuning data. Thus, we opted for a simpler approach where we start with a multilingual BERT model and fine-tune it on all the human ratings data available for all languages ($\X \rightarrow \Y$ and $\X\rightarrow \En$). In most cases, we found that such models could perform zero-shot evaluation: if a language $\Y$ does not have human ratings data, the metric can still perform evaluation in this target language as long as the base multilingual BERT model contains unlabeled data for~$\Y$, as observed in the past literature~\cite{karthikeyan2019cross, pires2019multilingual}.

We experiment with two pre-trained multilingual models: \textsc{mBERT} and \textsc{mBERT-WMT}, a custom multilingual variant of BERT. The \textsc{mBERT-WMT} model is larger that \textsc{mBERT} (24 Transformer layers instead of 12), and it was pre-trained on 19 languages of the WMT Metrics shared task 2015 to 2020. 

\paragraph{Details of \textsc{mBERT-WMT} pre-training} We trained  \textsc{mBERT-WMT} model with an MLM loss~\cite{devlin2018bert}, using a combination of public datasets: Wikipedia, the WMT 2019 News Crawl~\citep{barrault2019findings}, the C4 variant of Common Crawl~\citep{raffel2020exploring}, OPUS~\citep{tiedemann2012parallel}, Nunavut Hansard~\citep{joanis2020nunavut}, WikiTitles\footnote{\url{https://linguatools.org/tools/corpora/wikipedia-parallel-titles-corpora/}}, and ParaCrawl~\citep{espla2019paracrawl}. We trained a new WordPiece vocabulary~\citep{schuster2012_wpm,Wu2016_WPM}, since the original vocabulary of mBERT does not support the alphabets of Pashto, Khmer and Inuktitut. The model was trained for 1 million steps with the LAMB optimizer~\citep{You2019_LAMB}, using the learning rate 0.0018 and batch size 4096 on 64 TPU v3 chips.

\subsection{Experimental Setup}

\paragraph{Datasets} 
At the time of writing, no human ratings data is available for WMT Metrics 2020. Therefore, we use the human ratings from WMT Metrics years 2015 to 2019 for both training and evaluation. We do so in two stages. In the first stage, we use 2015 to 2018 for training (216,541~sentence pairs in 8 languages), setting 10\% aside for early stopping. We use 2019 as a development set, to choose hyper-parameters and to support high-level modeling decisions. In the second stage, we use 2015 to 2019, that is, all the data available, for training and uniformly sample 10\% of the data for 
early stopping and hyper-parameter tuning. This adds 289,895 sentence pairs and 4 additional languages to our training set, approximately doubling the size of the training data. We report our results on the first setup, but submit our predictions to the shared task using the second setup.

\paragraph{Hyper-parameters} We run grid search on the learning rate and export the best model, using values \{5e-6, 8e-6, 9e-6, 1e-5, 2e-5, 3e-5\}. We use batch size 32 and evaluate the model every 1,000 steps on a 10\% held-out data set to prevent over-fitting. During preliminary experiments, we additionally experimented with the batch size, dropout rate, frequency of continuous evaluation, balance of languages, pre-training schemes, WordPiece vocabularies, and model architecture.

\subsection{Additional Models and Baselines}

\paragraph{English BLEURT} We fine-tune a new \BLEURT{} checkpoint, following the methodology described above. The main difference with~\citet{sellam2020bleurt} is that we incorporate the to-English ratings of year 2019, which were not previously available.

\paragraph{Monolingual baselines based on BERT} We experiment with three baselines and submit the results to the WMT Metrics Shared Task for analysis. \textsc{BERT-L2-base} and \textsc{BERT-L2-large} are two regression models based on BERT and trained on to-English ratings. We use the same setup as English \BLEURT{}, but we omit the mid-training phase. A similar approach was described in~\citet{shimanaka2019machine}. \textsc{BERT-Chinese-L2} is similar to \textsc{BERT-L2-base}, but it uses \textsc{BERT-Chinese} and it is fine-tuned on to-Chinese ratings.

\paragraph{Other Systems} We compare our setups to other state-of-the-art learned metrics:  \textsc{BERTscore}~\citep{zhang2019BERTscore}, and Yisi~\citep{lo2019yisi} all apply rules on top of \textsc{BERT} embeddings while \textsc{ESIM}~\citep{mathur2019putting} is a neural sentence similarity model. \textsc{PRISM}~\citep{thompson2020automatic} trains a multilingual translation model that is used as a zero-shot paraphrasing system. All the aforementioned systems take sentences pairs as input. Concurrent work has investigated incorporating the source with great success~\cite{rei2020comet}. We leave this line of research for future work.

\section{Results}
\label{subsec:results}

\begin{table*}[ht]
\centering
\scriptsize
\begin{tabular}{@{}lcccccccc@{}}
  \toprule
     & de-en & fi-en & gu-en & kk-en & lt-en & ru-en & zh-en & \textbf{avg} \\
\midrule
\textsc{YiSi} & 0.164 & 0.347 & 0.312 & 0.440 & 0.376 & 0.217 & 0.426 & 0.326\\
\textsc{YiSi1-SRL} & 0.199 & 0.346 & 0.306 & 0.442 & 0.380 & 0.222 & 0.431 & 0.332 \\
\textsc{ESIM} & 0.167 & 0.337 & 0.303 & 0.435 & 0.359 & 0.201 & 0.396 & 0.314 \\
\textsc{BERTscore} & 0.176 & 0.345 & \textbf{0.320} & 0.432 & 0.381 & 0.223 & 0.430 & 0.330 \\
\textsc{PRISM} & \textbf{0.204} & 0.357 & 0.313 & 0.434 & 0.382 & 0.225 & 0.438 & 0.336 \\
\midrule
\textbf{BLEURT Configurations, English-only}\\
\textsc{BERT-L2-base} & 0.142 & 0.326 & 0.274 & 0.406 & 0.367 & 0.197 & 0.358 & 0.296 \\
\textsc{BERT-L2-large} & 0.172 & 0.361 & 0.305 & 0.424 & 0.388 & 0.210 & 0.420 & 0.326 \\
\textsc{BLEURT} & 0.175 & \textbf{0.365} & 0.316 & \textbf{0.451} & 0.397 & 0.223 & \textbf{0.444} & \textbf{0.339} \\
\midrule
\textbf{BLEURT Configurations, Multi-lingual}\\
\textsc{mBERT} & 0.172 & 0.352 & 0.300 & 0.430 & 0.388 & 0.222 & 0.397 & 0.323 \\
\textsc{mBERT}-WMT & 0.187 & 0.363 & 0.306 & 0.439 & \textbf{0.398} & \textbf{0.226} & 0.425 & 0.335 \\
\bottomrule
\end{tabular}
\caption{Segment-level agreement with human ratings on the WMT19 Metrics Shared Task on the to-English language pairs. The metric is WMT's Direct Assessment metric, a robust variant of Kendall $\tau$. The scores for \textsc{YiSi}, \textsc{YiSi1-SRL}, and \textsc{ESIM} come from~\newcite{ma2019results}. The scores for \textsc{BERTscore} and \textsc{PRISM} come from~\newcite{thompson2020automatic}.}
\label{table:wmt19en}
\end{table*}

\begin{table*}[ht]
\centering
\scriptsize
\begin{tabular}{@{}lcccccccc@{}}
  \toprule
     & de-en & fi-en & gu-en & kk-en & lt-en & ru-en & zh-en & \textbf{avg} \\
\midrule
\textsc{YiSi} & 0.949 & 0.989 & 0.924 & 0.994 & 0.981 & 0.979 & 0.979 & 0.971 \\
\textsc{YiSi1-SRL} & 0.950 & 0.989 & 0.918 & 0.994 & 0.983 & 0.978 & 0.977 & 0.969 \\
\textsc{ESIM} & 0.941 & 0.971 & 0.885 & 0.986 & 0.989 & 0.968 & 0.988 & 0.961 \\
\textsc{BERTscore} & 0.949 & 0.987 & 0.981 & 0.980 & 0.962 & 0.921 & 0.983 & 0.966\\
\textsc{PRISM} & \textbf{0.954} & 0.983 & 0.764 & \textbf{0.998} & \textbf{0.995} & 0.914 & 0.992 & 0.943\\
\midrule
\textbf{BLEURT Configurations, English-only}\\
\textsc{BERT-L2-base} & 0.938 & \textbf{0.992} & \textbf{0.930} & 0.992 & 0.991 & 0.976 & \textbf{0.997} & \textbf{0.974}\\
\textsc{BERT-L2-large} & 0.940 & 0.987 & 0.819 & 0.992 & 0.990 & \textbf{0.985} & 0.993 & 0.958\\
\textsc{BLEURT} & 0.943 & 0.989 & 0.865 & 0.996 & \textbf{0.995} & 0.984 & 0.990 & 0.966\\
\midrule
\textbf{BLEURT Configurations, Multi-lingual}\\
\textsc{mBERT} & 0.937 & 0.976 & 0.863 & 0.984 & 0.978 & 0.959 & 0.978 & 0.954\\
\textsc{mBERT-WMT} & 0.950 & 0.991 & 0.815 & 0.989 & 0.992 & 0.968 & 0.980 & 0.955\\
\bottomrule
\end{tabular}
\caption{System-level agreement with human ratings on the WMT19 Metrics Shared Task on the to-English language pairs. The metric is Pearson's correlation.  The scores for \textsc{YiSi}, \textsc{YiSi1-SRL}, and \textsc{ESIM} come from~\newcite{ma2019results}. The scores for \textsc{BERTscore} and \textsc{PRISM} come from~\newcite{thompson2020automatic}.}
\label{table:wmt19en-sys}
\end{table*}

\begin{table*}[ht]
\centering
\scriptsize
\begin{tabular}{@{}lcccccccccccc@{}}
  \toprule
     & en-cs & en-de & en-fi & \textit{en-gu} & \textit{en-kk} & \textit{en-lt} & en-ru & en-zh & de-cs & \textit{de-fr} & fr-de & \textbf{avg} \\ 
\midrule
\textsc{YiSi1}  & 0.475 & 0.351 & 0.537 & 0.551 & 0.546 & 0.470 & 0.585 & 0.355 & 0.376 & 0.349 & 0.310 & 0.446 \\
\textsc{YiSi1-SRL} & - & 0.368 & - & - & - & - & - & 0.361 & - & - & 0.299 & - \\
\textsc{ESIM} & - & 0.329 & 0.511 & - & 0.510 & 0.428 & 0.572 & 0.339 & 0.331 & 0.290 & 0.289 & - \\
\textsc{BERTscore} & 0.485 & 0.345 & 0.524 & 0.558 & 0.533 & 0.463 & 0.580 & 0.347 & 0.352 & 0.325 & 0.274 & 0.435 \\
\textsc{PRISM} & 0.582 & \textbf{0.426} & 0.591 & 0.313 & 0.531 & 0.558 & 0.584 & 0.376 & 0.458 & \textbf{0.453} & 0.426 & 0.482 \\
\midrule
\textbf{\BLEURT{} Configurations}  &&&&&&&&&&&& \\
\textsc{BERT-Chinese-L2} & - & - & - & - & - & - & - & 0.356 & - & - & - & -\\
\textsc{mBERT} & 0.506 & 0.364 & 0.551 & 0.550 & 0.529 & 0.516 & \textbf{0.592} & \textbf{0.381} & 0.385 & 0.388 & 0.291 & 0.459\\
\textsc{mBERT-WMT} & \textbf{0.603} & 0.422 & \textbf{0.615} & \textbf{0.577} & \textbf{0.558} & \textbf{0.584} & 0.492 & 0.337 & \textbf{0.461} & 0.449 & \textbf{0.427} & \textbf{0.502}\\
\bottomrule
\end{tabular}
\caption{Segment-level agreement with human ratings on the WMT19 Metrics Shared Task on non-English language pairs. The metric is WMT's Direct Assessment metric, a robust variant of Kendall $\tau$. Languages without fine-tuning data are denoted in \textit{italics}.  The scores for \textsc{YiSi}, \textsc{YiSi1-SRL}, and \textsc{ESIM} come from~\newcite{ma2019results}. The scores for \textsc{BERTscore} and \textsc{PRISM} come from~\newcite{thompson2020automatic}.}
\label{table:wmt19multi}
\end{table*}

\begin{table*}[ht]
\centering
\scriptsize
\begin{tabular}{@{}lcccccccccccc@{}}
  \toprule
     & en-cs & en-de & en-fi & \textit{en-gu} & \textit{en-kk} & \textit{en-lt} & en-ru & en-zh & de-cs & \textit{de-fr} & fr-de & \textbf{avg} \\ 
\midrule
\textsc{YiSi1} & 0.962 & \textbf{0.991} & 0.971 & 0.909 & 0.985 & 0.963 & \textbf{0.992} & 0.951 & 0.973 & 0.969 & 0.908 & 0.961\\
\textsc{YiSi1-SRL} & - &\textbf{0.991} & - & - & - & - & - & 0.948 & - & - & 0.912 & - \\
\textsc{ESIM} & - & \textbf{0.991} & 0.957 & - & 0.980 & \textbf{0.989} & 0.989 & 0.931 & 0.980 & 0.950 & 0.942 & - \\
\textsc{BERTscore} & 0.981 & 0.990 & 0.970 & 0.922 & 0.981 & 0.978 & 0.989 & 0.925 & 0.969 & 0.971 & 0.899 & 0.961\\
\textsc{PRISM} & 0.958 & 0.988 & 0.949 & 0.624 & 0.978 & 0.937 & 0.918 & 0.898 & 0.976 & 0.936 & 0.911 & 0.916 \\
\midrule
\textbf{\BLEURT{} Configurations}  &&&&&&&&&&&& \\

\textsc{BERT-Chinese-L2} & - & - & - & - & - & - & - & \textbf{0.953} & - & - & - & -\\
\textsc{mBERT} & 0.942 & 0.987 & 0.953 & 0.949 & 0.982 & 0.950 & 0.947 & 0.949 & 0.972 & 0.970 & 0.924 & 0.957\\
\textsc{mBERT-WMT} & \textbf{0.993} & \textbf{0.991} & \textbf{0.987} & \textbf{0.959} & \textbf{0.993} & \textbf{0.989} & 0.888 & \textbf{0.953} & \textbf{0.986} & \textbf{0.988} & \textbf{0.962} & \textbf{0.972}\\
\bottomrule
\end{tabular}
\caption{System-level agreement with human ratings on the WMT19 Metrics Shared Task on non-English language pairs. The metric is Pearson's correlation. Languages without finetuning data are denoted in \textit{italics}.  The scores for \textsc{YiSi}, \textsc{YiSi1-SRL}, and \textsc{ESIM} come from~\newcite{ma2019results}. The scores for \textsc{BERTscore} and \textsc{PRISM} come from~\newcite{thompson2020automatic}.}
\label{table:wmt19multi-sys}
\end{table*}

Tables~\ref{table:wmt19en} and~\ref{table:wmt19en-sys} show the results in the $\X \rightarrow \En$ direction, at the segment- and system-level respectively. In the majority of cases, one of the \BLEURT{} configurations yields the strongest results. The original \BLEURT{} metric seems to perform better at the segment-level. At the system-level it may be dominated by PRISM (3 out of 7 language pairs) or by one of the simpler BERT-based models (4 out of 7 language pairs).

Tables~\ref{table:wmt19multi} and~\ref{table:wmt19multi-sys} present the results for the other languages. \textsc{mBERT-WMT} yields solid results at the segment-level (it achieves the highest correlations for 7 out of 11 language pairs), in particular for the ``zero-shot'' setups, $\En \rightarrow \Gu$, $\En \rightarrow \Kk$, and $\En \rightarrow \Lt$. It outperforms \textsc{mBERT} consistently, except for $\En \rightarrow \Ru$ and $\En \rightarrow \Zh$ where it lags behind the other metrics. The results are consistent at the system-level.

\paragraph{Strategy for the WMT Metrics Shared Task} Based on these results, we make two ``competitive'' submissions. We present \BLEURT{} as described above, which we ran on all the $\X \rightarrow \En$ sentence pairs. Additionally, we submitted a multilingual system that combines \textsc{mBERT-WMT} (for all languages except Chinese) and \textsc{BERT-Chinese-L2} (for Chinese). We ran the multilingual system for all language pairs including to-English, as the large amount of non-English fine-tuning data made available in 2019 may benefit this setup too. We also release the predictions of \textsc{BERT-Base-L2}, \textsc{BERT-Large-L2}, and \textsc{mBERT} for analysis.

\section{Additional Improvements on English$\to$German}
\label{subsec:ende}
For English$\to$German, the organizers of WMT20 provide three different reference translations: two standard references and one additional paraphrased reference. Given this novel setup, we investigate how to combine our predictions. Moreover, we use a similar framework to ensemble the predictions of different metrics. In particular, we average the predictions of \textsc{BLEURT}, \textsc{YiSi-1} and \textsc{YiSi-2}. All three metrics are different in their approaches. While \textsc{BLEURT} and \textsc{YiSi-1} are reference-based metrics, \textsc{Yisi-2} is reference-free and calculates its score by comparing translations only to the source sentence. \textsc{BLEURT} is fine-tuned on previous human ratings, while \textsc{YiSi-1} is based on the cosine similarity between \textsc{BERT} embeddings of the reference and the candidate. 

In the remainder of this section, we report \BLEURT{} results using the \textsc{mBERT-WMT} setup unless specified otherwise.\footnote{We use a different checkpoint from the one described in Section~\ref{subsec:results}. The model was trained for 880K steps instead of 1 million, and it uses a sequence length of 256 tokens instead of 128.}

\subsection{Modifications to YiSi-1}
Before combining \textsc{BLEURT} and \textsc{YiSi}, we perform a series of modifications to \textsc{YiSi-1} and evaluate their impact on English$\to$German.

\paragraph{Experimental Setup}
All experimental results are summarized in Table~\ref{table:wmt19ende}. We report both segment-level (DARR) and system-level (Kendall~$\tau$) correlations. To replicate the multi-reference setup of 2020, we compute correlations with the standard WMT references as well as the paraphrased reference from \newcite{freitag2020bleu}. 

\paragraph{Improving YiSi's Predictions} Our baseline is similar to the \textsc{YiSi-1} submission from WMT 2019~\cite{lo2019yisi}: we run \textsc{YiSi-1} with the public multilingual \emph{\textsc{mBERT}} checkpoint.
We then experiment with the underlying checkpoint. We continued pre-training \textsc{mBERT} on the in-domain German NewsCrawl dataset. The resulting model \emph{+pre-train NewsCrawl layer 9} increases the correlation for both reference translations. We  improve the correlation further on the paraphrased reference  by using the 8th instead of the 9th layer.

\paragraph{Other experiments} We tried pre-training BERT on forward translated sentences from German NewsCrawl, to adapt the word embeddings to MT outputs. We also trained a BERT model from scratch on the German NewsCrawl data. These experiments did not result in higher correlations with human ratings.

\begin{table}[ht]
\centering
\scriptsize
{\setlength{\tabcolsep}{.4em}
\begin{tabular}{@{}lllcc@{}}
  \toprule
        &  &  & sys-level & seg-level  \\ 
    Ref & Metric & model & Kendall~$\tau$ & DARR  \\ 
\midrule
    std & BLEURT & \textsc{mBERT}-WMT$^\mathparagraph$ &  \bf{0.896} & \bf{0.420} \\ \hline
    \multirow{3}{*}{std} & \multirow{3}{*}{YiSi-1} & \textsc{mBERT} \emph{(WMT19 subm.)} & 0.810 & 0.351 \\
    & & \ +pre-train NewsCrawl layer 9 & 0.870 & 0.373 \\
    & & \ +pre-train NewsCrawl layer 8~$^\dagger$ & 0.853 & 0.376 \\
    \hline \hline
     para & BLEURT & \textsc{mBERT}-WMT$^\mathparagraph$ & 0.852 & \bf{0.413} \\ \hline 
    \multirow{3}{*}{para} &  \multirow{3}{*}{YiSi-1} & \textsc{mBERT} \emph{(WMT19 subm.)} & 0.844 & 0.316 \\
    & & \ +pre-train NewsCrawl layer 9 & 0.887 & 0.365 \\
    & & \ +pre-train NewsCrawl layer 8~$^\dagger$ & \bf{0.896} & 0.373 \\
    \hline \hline
    src & YiSi-2 & \textsc{mBERT}$^\mathparagraph$ & 0.307 & 0.106 \\
    \hline \hline
    \multirow{2}{*}{2std+para} & YiSi-comb & comb of 3 ($\dagger$ systems)  & \bf{0.905} & 0.399 \\
     & all-comb & avg of 7 ($\dagger$ \& $\mathparagraph$ systems) & 0.878 & \bf{0.454} \\
\bottomrule
\end{tabular}
}
\caption{Agreement with human ratings on the WMT19 Metrics Shared Task for English$\to$German. The first set of results are generated by using the standard reference translations for WMT 2019. The second set of results is generated by using the paraphrased reference translations. YiSi-2 is reference free and only uses the source sentences.}
\label{table:wmt19ende}
\end{table}

\subsection{Combining \BLEURT{}, \textsc{YiSi-1} and \textsc{YiSi-2} on Multiple References}

We describe our two submissions to WMT 2020, \textsc{YiSi-comb} and \textsc{All-comb}, which result from our efforts to use multiple references for automatic evaluation. \textsc{YiSi-comb} is a multi-reference version of the \textsc{YiSi} score~\cite{lo2019yisi} aimed at achieving better system-level correlations. \textsc{All-comb} leverages metrics from \BLEURT{}, \textsc{YiSi-1}, and \textsc{YiSi-2} on multiple references to achieve better segment-level correlation.

\begin{figure}
    \centering
    \includegraphics[width = \linewidth]{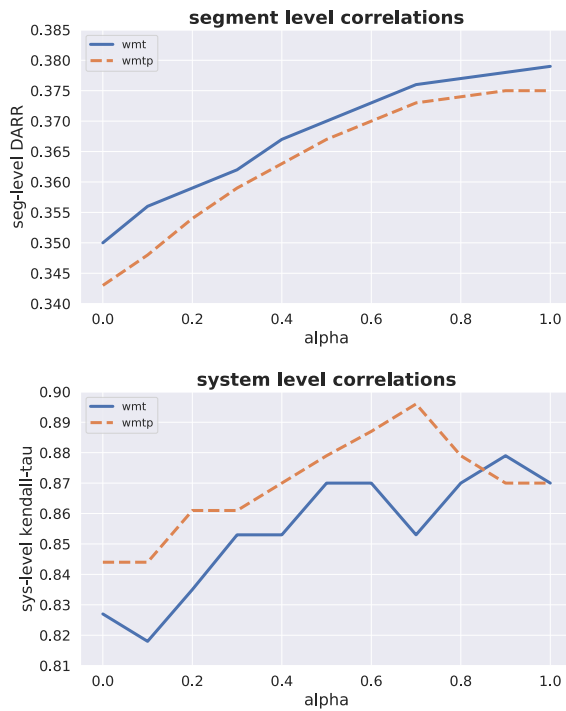}
    \caption{Correlations with respect to different $\alpha$ settings for Yisi-1. The system-level correlation is highest when $\alpha = 0.7$, which is the $\alpha$ we use for the submission.}
    \label{fig:wmt2019_alpha}
\end{figure}

\paragraph{\textsc{YiSi-comb}}
\textsc{YiSi} scores are $F_1$ scores of \textsc{YiSi} precision and \textsc{YiSi} recall. For the \textsc{YiSi-comb} submission, we take the minimum of the \textsc{YiSi} recalls for the three different references as the multi-reference recall, and the maximum of the \textsc{YiSi} precision as the multi-reference precision. Using the same notations as in \citep{lo2019yisi}, the final score is the $F_1$ of the recall and precision computed with $\alpha~\textrm{=}~0.7$ (see Figure~\ref{fig:wmt2019_alpha}). This submission aims to maximize the system-level correlation.

As shown in Table~\ref{table:wmt19ende}, \textsc{YiSi-1} has the highest system-level correlation on paraphrased references. Given that we used $\alpha~\textrm{=}~0.7$, \textsc{YiSi} scores are quite similar to \textsc{YiSi} recalls (when $\alpha~\textrm{=}~1.0$, \textsc{YiSi} scores are equal to \textsc{YiSi} recalls). \textsc{YiSi-1} scores for paraphrased references are usually much lower than those of standard references, therefore taking the minimum recall is oftentimes equivalent to taking the \textsc{YiSi} recall from the paraphrased references. Furthermore, we found that using the maximum precision, in combination with aggregating recalls, usually performs the best.

\paragraph{\textsc{All-comb}}
We combined the predictions of \textsc{YiSi-1} with those of \BLEURT{} and \textsc{YiSi-2}. \textsc{YiSi-2} usually performs worse than the reference-based metrics, but we found that incorporating its predictions can help. 
Having three different metrics (\BLEURT{}, \textsc{YiSi-1}, \textsc{YiSi-2}) and three different reference translations, we take all seven predictions and average the scores for each segment. The combined prediction \textsc{All-comb} outperforms every single metric at the segment level, though the system-level correlation drops in comparison to the best \textsc{YiSi-1} score on paraphrased references. This submission aims to maximize the segment-level correlation.

\section{Summary}
We submit the following systems to the WMT Metrics shared task:
\begin{itemize}
\item \BLEURT{} as previously published, fine-tuned on the human ratings of the WMT Metrics shared task 2015 to 2019, to-English.
\item A multi-lingual extensions of \BLEURT{} based on a 20 languages variant of \textsc{mBERT} and \textsc{BERT-Chinese}.
\item Three baseline systems based on \textsc{BERT-base}, \textsc{BERT-large}, and \textsc{mBERT}.
\item Two combination methods for English to German that use YiSi and alternative references, \textsc{YiSi-comb} and \textsc{All-comb}.
\end{itemize}

\section{Acknowledgements}
Thanks to Xavier Garcia and Ran Tian for advice and proof-reading.
\bibliography{main}
\bibliographystyle{acl_natbib}

\end{document}